\documentclass[letterpaper, 10 pt, conference]{ieeeconf}

\IEEEoverridecommandlockouts

\usepackage{xcolor}
\usepackage{amssymb}
\usepackage{graphicx}
\usepackage{mathtools}
\usepackage{amsmath}
\usepackage{gensymb}
\usepackage{float}
\usepackage{multirow}
\usepackage{makecell}
\usepackage{mathtools}
\usepackage{hyperref}
\usepackage{overpic}
\usepackage{subcaption}
\usepackage{xcolor}
\hypersetup{
    colorlinks=true,
    linkcolor=blue,
    filecolor=magenta,      
    urlcolor=blue,
    pdftitle={Overleaf Example},
    pdfpagemode=FullScreen,
    }

\usepackage{tabularx}
    \newcolumntype{L}{>{\raggedright\arraybackslash}X}
\usepackage[utf8]{inputenc}
\usepackage[english]{babel}

\usepackage{amsthm}
\usepackage{booktabs}
\usepackage{algorithm}
\usepackage{algpseudocode}

\usepackage{subcaption}

\newcommand{\ours}{AMCO }

\newcommand{\firstletter}{g }
\newcommand{\secondletter}{h }
\newcommand{\thirdletter}{p }
\newcommand{\finalletter}{coupled }

\newcommand{\no}{\noindent}

\newcommand{\firstmap}{general knowledge map}
\newcommand{\secondmap}{history map}
\newcommand{\thirdmap}{proprioception map}
\newcommand{\finalmap}{coupled map}

\title{\LARGE \bf  AMCO:  Adaptive Multimodal Coupling of Vision and Proprioception for Quadruped Robot Navigation in Outdoor Environments
\\}

\author{Mohamed Elnoor, Kasun Weerakoon, Adarsh Jagan Sathyamoorthy, Tianrui Guan, Vignesh Rajagopal, \\and Dinesh Manocha \\ {\small Technical report and video can be found at \url{http://gamma.umd.edu/amco/}}}

\begin{document}

\maketitle

\begin{abstract}
We present AMCO, a novel navigation method for quadruped robots that adaptively combines vision-based and proprioception-based perception capabilities. Our approach uses three cost maps: general knowledge map; traversability history map; and current proprioception map; which are derived from a robot's vision and proprioception data, and couples them to obtain a coupled traversability cost map for navigation. The general knowledge map encodes terrains semantically segmented from visual sensing, and represents a terrain's typically expected traversability. The traversability history map encodes the robot's recent proprioceptive measurements on a terrain and its semantic segmentation as a cost map. Further, the robot's present proprioceptive measurement is encoded as a cost map in the current proprioception map. As the general knowledge map and traversability history map rely on semantic segmentation, we evaluate the reliability of the visual sensory data by estimating the brightness and motion blur of input RGB images and accordingly combine the three cost maps to obtain the coupled traversability cost map used for navigation. Leveraging this adaptive coupling, the robot can depend on the most reliable input modality available. Finally, we present a novel planner that selects appropriate gaits and velocities for traversing challenging outdoor environments using the coupled traversability cost map. We demonstrate AMCO's navigation performance in different real-world outdoor environments and observe 10.8\%-34.9\% reduction w.r.t. two stability metrics, and up to 50\% improvement in terms of success rate compared to current navigation methods.

\end{abstract}


%
%

\section{Introduction}  \label{sec:Intro}


Quadruped robots have been used for autonomous navigation due to their superior dynamics, which enables them to traverse diverse and challenging terrains in outdoor settings \cite{frey2023fast, sathyamoorthy2023vern}. They can be used in applications including agriculture \cite{sotnik2022agricultural}, planetary exploration \cite{valsecchi2023towards}, surveillance \cite{bruzzone2016functional}, and disaster response missions \cite{yoshiike2017development}. However, guiding these robots across such complex environments requires a comprehensive understanding of the scene, which encompasses 
how the terrain visually appears using vision-based sensors and how it physically \textit{feels} under the robot's feet using proprioception.

Traditionally, terrain traversability has been assessed using exteroceptive/visual sensors such as cameras \cite{sun2017vision,guan2022ga} and LiDARs \cite{li2023seeing,sathyamoorthy2023using}, which provide valuable information about the environment. Yet, these approaches often fall short in complex terrains where visual features do not provide sufficient information about the terrain, or when the visual characterization is incorrect. For instance, a terrain that appears to be solid and stable (like soil) but deforms (is actually mud) under the robot's weight. This leaves the robot vulnerable to sinkages, slips, and entrapments \cite{valsecchi2023towards,dey2022prepare, weerakoon2023vapor}. 

On the other hand, proprioception, which can be captured from a legged robot's internal joint encoders \cite{fu2022coupling, elnoor2023pronav}, odometry errors \cite{frey2023fast}, robot states \cite{yao2022rca},and contact vibrations \cite{brooks2007self}, effectively measures the terrain traversability characteristics (softness, bumpiness, etc) at the robot's present location \cite{al2020review}. Proprioception has previously been used to learn traversability costs for visual navigation \cite{frey2023fast,yao2022rca}. However, it can not be used to predict the terrain traversability ahead of the robot \cite{elnoor2023pronav} without exteroceptive sensing.



\begin{figure}[t]
    \centering
    \includegraphics[width=0.9\columnwidth]{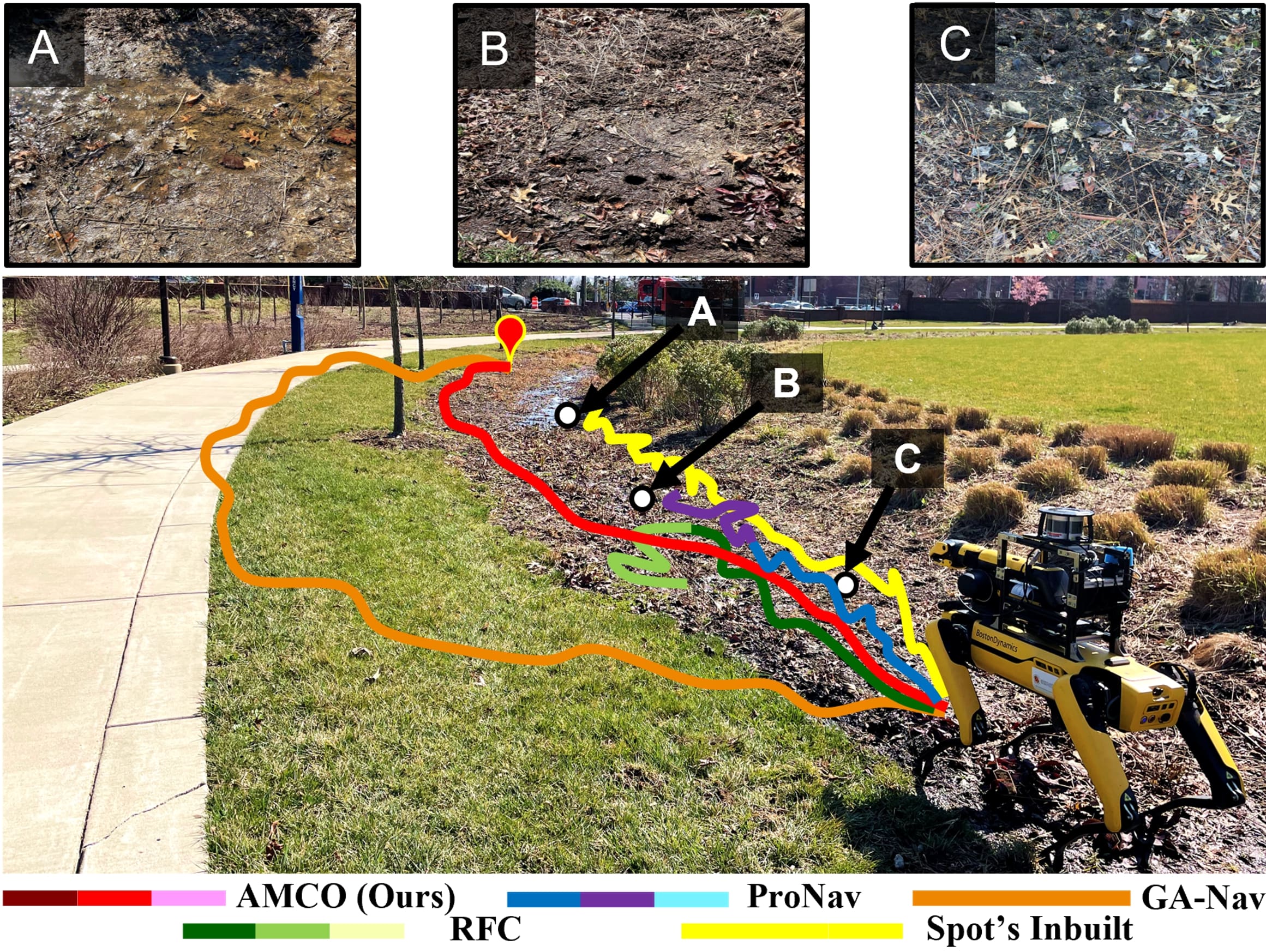}
    \caption{\small{The navigation trajectories generated by our method, AMCO (trot: brown, crawl: red, amble: pink), ProNav \cite{elnoor2023pronav} (trot: blue, crawl: purple, amble: light blue), GaNav \cite{guan2022ga} (orange), Spot's Inbuilt (yellow), and RFC \cite{kertesz2016rigidity} (trot: green, crawl: light green, amble: light beige) in scenario 4, which contains different types of deformable terrains: (A) water puddle, (B) mud, (C) granular. AMCO, ProNav, and RFC can switch between three gaits (trot, crawl, and amble) during navigation as presented in three different colors. 
    Our method AMCO updates its trajectory as traversing the terrain and uses the coupled vision and proprioception to avoid extremely deformable regions such as A and B. It balances between the trajectory length and maintaining stability to reduce potential robot failures on terrains with varying deformability. Conversely, GaNav \cite{guan2022ga} takes a longer path by avoiding mud and grass altogether. The other methods takes a straight path towards the goal which leads to sinkage and failures. 
   }}
    \label{fig:cover}
    \vspace{-15pt}
\end{figure}


To this end, many approaches have combined proprioceptive and visual inputs using self-supervised learning \cite{frey2023fast,fu2022coupling,yao2022rca,wisth2022vilens,jung2023v}. Those methods estimate traversability cost maps which only represent terrains' \textit{typical} traversability. However, they do not account for the potential changes in terrains' physical properties, such as deformability \cite{dallas2021terrain} and trafficability \cite{comin2017trafficability},
due to weather or other factors. This limitation restricts previous methods from adapting to traversability changes, potentially leading to failures. To create a generalizable traversability estimation approach for stable navigation, the robot must also consider both the recent proprioceptive history, and the current instant's proprioceptive feedback from the terrain along with its \textit{typical} traversability obtained from visual inputs.

\textbf{Main Contributions:} 
To this end, we present \ours (\textbf{A}daptive \textbf{M}ultimodal \textbf{CO}upling of vision and proprioception for stable robot navigation), a novel approach that adaptively integrates semantic segmentation and proprioceptive feedback of terrain into a traversability cost map. This multi-modal perception representation is integrated with a local planner to select a stable gait and velocity combination for robot navigation. 
Our approach is inspired by how humans navigate through deformable, and occluded terrains like sand, muddy grass, and bushes, using: 1. our general knowledge of the terrain (e.g. \textit{soil is usually stable to walk on}), 2. recent traversability history (e.g. \textit{soil was wet and felt deformable a few time steps ago in a nearby location}), 3. current experience (e.g. \textit{soil feels more deformable and non-traversable}). Here, both visual features and proprioception information inform our understanding of the terrain and guide our movement. The novel components of our work are:

\begin{itemize}
  \item We introduce novel formulations of three traversability cost maps using vision and proprioception perception capabilities. 
  A \firstmap \, that uses semantic segmentation from an RGB camera to represent the terrain's traversability ahead of the robot. Next, we use the recent history of proprioception signals, which are measured from the robot's joint positions, forces, and current consumption from the battery. 
  We captured the recent duration of this prospective-informed traversability and its consistency with its semantic traversability class to
  to build a history map.
  Also, we present a proprioceptive-based traversability cost map, which extrapolates the current robot proprioception to future time steps to robustly predict the traversability along the robot's trajectory.

  \item An adaptive multi-modal coupling of the three cost maps according to the vision-based sensor reliability. 
  To do that, we incorporate a reliability estimation method that assesses the quality of RGB images and weighs the vision-based representations on the \finalmap traversability cost map. This ensures that the navigation decisions are based on the most reliable input modality available. We demonstrate that our adaptive coupling leads to improvements in navigating different types of deformable terrains.
  
  \item A navigation algorithm that adapts gait selection and dynamically calculates feasible velocities in real-time based on multi-modal sensory input (\finalmap)  to ensure stable navigation. We demonstrate our method on Boston dynamic spot and evaluate its performance in four unstructured outdoor environments. We observe 10.8\%-34.9\% reduction in IMU Energy density \cite{try2023vibration} and up to 50\% improvement in navigation success rate.

\end{itemize}


\section{Related Works}
In this section, we discuss the traversability estimation methods for both exteroception and proprioception sensors. After that, we briefly discuss the existing methods for outdoor navigation.

\subsection{Traversability Estimation}
Traversability estimation tends to incorporate vision-based approaches that use external sensors and proprioception-based approaches. 
We review both perception methods in the following sections.

\subsubsection{Exteropercption-based perception}


Exteroception has been widely used to estimate traversability in outdoor settings \cite{sathyamoorthy2023using, al2020low,yasuda2020autonomous}. Many methods leverage semantic segmentation to classify different types of terrains in outdoor settings \cite{guan2022ga,miyamoto2020visual,zhan2020adaptive}. For instance, ViTAL \cite{fahmi2022vital} leverages vision-based techniques for terrain-aware locomotion, which optimizes path planning across varied terrains with a focus on quadured robots. Another noteworthy approach by Agarwal et al.\cite{agarwal2022legged}, utilizes an egocentric camera to facilitate legged locomotion on rugged terrains through an end-to-end locomotion system. These methodologies predominantly utilize vision-based sensors, 
whose performance can degrade under various conditions such as light changes, motion blur, and occlusions \cite{aladem2019evaluation}.\\


\subsubsection{Proprioception-based Perception}

Proprioception, the internal sense of movement and position, plays a crucial role in quadruped robot navigation, as described in \cite{frey2023fast,fu2022coupling,haresh2023learning}.  It is defined as the robot's internal sense of movement and position \cite{al2020review}. While effective in terrain interaction, its limitations in predicting unencountered terrain necessitate external sensor integration for improved navigation. Wisth et al.'s development of a method combining proprioceptive feedback with visual, inertial, lidar, and odometry data exemplifies advancements in multi-modal sensor fusion \cite{wisth2022vilens}. This approach, further refined by cross-modal learning techniques by Loquercio et al. \cite{loquercio2023learning}, enhances environmental understanding and adaptability. Moreover, studies by García et al. \cite{garcia2022gait} and Karnan et al. \cite{haresh2023learning,karnan2023sterling} illustrate the benefits of using proprioceptive data for gait adjustments and incorporating human preferences into navigation systems, showcasing the potential of learning-based methods for better terrain adaptation. This collective research shows the significance of integrating proprioceptive feedback with external sensors for advanced robotic navigation and terrain adaptability.

Other works include fused proprioception with visual data in particular to improve navigation. As highlighted by Fu et al. \cite{fu2022coupling}, integrating these sensory inputs improves spatial awareness and movement strategies. This approach is crucial for navigating adverse conditions, with Teng et al. \cite{teng2021legged} and Jin et al. \cite{jin2023resilient} showcasing the importance of proprioception in maintaining balance and enhancing resilience. Assessing terrain supportiveness through proprioceptive data is essential for safe and effective navigation. Homberger et al. \cite{homberger2019support} discuss the role of proprioception in evaluating surface stability. Recently, Dey et al. \cite{dey2022prepare} introduced a predictive method that utilizes proprioceptive signals to foresee and counteract potential navigation failures, emphasizing the need for anticipation in challenging terrains.
While these methods integrate proprioception with exteroception, many of them
assume the consistency between the two sensory modalities which is reasonable for some terrains. 
In our case, we
consider terrains that could have different traversability under different weather conditions such as sand, mud, grass,
and covered grounds.

\subsection{Outdoor Navigation}

Recent advancements in outdoor navigation for autonomous systems have showcased the integration of several methods and hardware designs \cite{valsecchi2023towards, fahmi2022vital, weerakoon2023adventr, dwa}. For instance, \cite{overbye2021path} works on path optimization for ground vehicles in off-road terrains uses vehicle kinematic constraints and advanced algorithms for improved efficiency. On the other hand, \cite{wellhausen2021artplanner} developed robust navigation systems for legged robots, employing reachability-based methods and neural network-based learned motion costs to navigate unpredictable terrains in term of traversability. Miki et al. \cite{miki2022learning} advanced sensory integration techniques, employing attention-based recurrent encoders and lidar intensity maps for enhanced environmental interaction.
Additionally, \cite{valsecchi2023towards} investigated the potential of quadruped robots for haptic inspection of planetary soils. 
Also, Kolvenbach et al. in \cite{kolvenbach2019haptic} have developed a terrain-aware locomotion controller. This controller combines proprioceptive and exteroceptive inputs using a recurrent encoder, which enables the legged robots to navigate challenging terrains. Our navigation approach is different from these methods. We compute the best gait and velocity pairs based on the semantic traversability class and proprioception. This helps the robot to maintain stability as well as heading toward the goal.


\section{Background}
In this section, we explain symbols, notations, and the preliminary concepts used in our work. 
\subsection{Symbols and Notations}
We highlight the symbols of images and cost maps we use in Table (\ref{tab:symbols}).
\begin{table}[h]

\label{tab:symbols}
\centering
\small
\begin{tabular}{|c|l|}
\hline
\textbf{Symbols} & \textbf{Definitions} \\
\hline
\( I_{RGB}^t \) & \scriptsize Input RGB image. \\ [0.7ex]
\hline
\( I_{SD}^t \) & \scriptsize Image after applying semantic segmentation. \\[0.6ex]
\hline
\( \mathcal{C}_{\firstletter}^t \) &  \scriptsize A \firstmap \, based on vision . \ref{sec:our approach-prior} \\[0.7ex]
\hline
\( \mathcal{C}_{\secondletter}^t \) & \scriptsize A \secondmap\,that's based on vision and proprioception. \ref{sec:our approach-history}\\[0.7ex]
\hline
\( \mathcal{C}_{\thirdletter}^t \) &  \scriptsize A \thirdmap\,that's based on proprioception. \ref{sec:our approach-proprioception}\\[0.7ex]
\hline
\( \mathcal{C}_{\finalletter}^t \) & \scriptsize A \finalmap\ that's computed after the adaptive coupling. \ref{sec:adaptive-coupling} \\[0.7ex]

\hline
\end{tabular}
\caption{ \small{List of symbols used in our approach.  $ I_{RGB}^t$ and $ I_{S}^t$ are three-channel images of size $H \times W$. All the cost maps are grayscale of size $H \times W$ where $H, W$ are the height and width of the images and the cost maps.}  }
\label{tab:symbols}
\vspace{-10pt}
\end{table}

\subsection{Setup and Conventions} \label{sec:setup-conventions}
The quadruped robot in our formulation has a coordinate frame attached to its center of mass with X, Y, and Z directions facing forward, leftward, and upward, respectively. The robot is equipped with sensors to measure the positions, 
velocities and forces exerted at each joint in the $i^{th}$ leg along each axis. Additionally, the robot measures the current $\mathcal{I}^t$ consumed from the battery. At every time instant $t$, these quantities are used as raw proprioceptive signals to \textit{feel} the terrain and estimate its traversability (section \ref{sec:pronav-background}). The robot is also equipped with an RGB camera that provides an image $I_{RGB}^t$ to view the terrains ahead, and a 3D lidar to detect obstacles. The robot can execute three kinds of gaits: \textbf{1. trot}, where two of the robot's feet make contact with the ground while walking, \textbf{2. crawl}, where three feet touch the ground for enhanced stability but with a restricted maximum velocity, and \textbf{3. amble}: a gait similar to crawl but with unrestricted maximum speed. 

We categorize the various terrains that the robot navigates into the following types: 1. stable, 2. granular (e.g. sand, snow), 3. poor foothold (e.g. rocks), 4. high resistance (e.g. dense vegetation). We use $i, j$ to indicate grid locations in costmaps.


\subsection{Proprioception-based Terrain Analysis} \label{sec:pronav-background}
\begin{figure}[h]
\centering
\includegraphics[width=0.9\columnwidth]{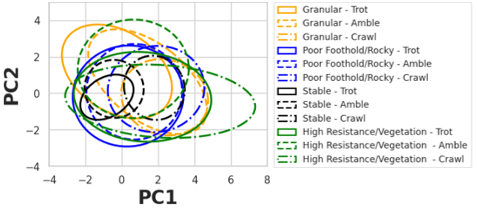}
\caption{\small{ The gaussian ellipses for the PCA components of four different terrains \cite{elnoor2023pronav}. For each terrain, three types of gait data are shown using different ellipse boundaries. The solid line denotes trot, the dashed line denotes amble, and the dash-dot line is for crawl.}}
\label{fig:ellipses}
\vspace{-15pt}
\end{figure}

We utilize the approach proposed in \cite{elnoor2023pronav} to estimate a terrain $\tau$'s traversability by processing the raw proprioceptive signals (defined in Section \ref{sec:setup-conventions}) from a legged robot and transforming them into two dimensional point $\mathbf{p}^t_{\tau}$ using Principal Component Analysis \cite{mackiewicz1993principal}. \cite{elnoor2023pronav} shows that the L-2 norm ($|\mathbf{p}^t_{\tau}|_2$) indicates a terrain's traversability. After estimating a terrain's traversability,  \cite{elnoor2023pronav} empirically demonstrates that using a certain gait $g$ to traverse a terrain type $\tau$ (\ref{sec:setup-conventions}) results in a bounded distribution of points in the PCA space, that can be modeled as a 2D Gaussian ellipse $\mathcal{E}(\tau, g)$. \cite{elnoor2023pronav} also shows that the area of a gait-terrain ellipse is directly proportional to the level of stability the robot experiences while traversing the terrain using the gait (see Fig. \ref{fig:ellipses}).

\subsection{Vision-based Traversability Types} \label{sec:segmentation model-background}

As a preliminary for our vision module, we use a terrain semantic segmentation method to classify the terrain type. 
Particularly, we incorporate \cite{guan2022ga}, a semantic segmentation method that uses group-wise attention to differentiate the traversability of the terrains and provides a segmentation image. We train this segmentation method using RUGD dataset \cite{RUGD2019IROS} using 4 terrain groups as presented in Table \ref{tab:terrain_types}.

\begin{table}[h]
\centering
\begin{tabular}{|c|l|}
\hline
\textbf{Semantic Categories} & \textbf{Terrain Type} \\ 
\hline
Concrete, Asphalt & Stable \\
\hline
Sand, Dirt, Gravel, Mulch & Granular \\
\hline
Rocks, Rockbed & Poor Foothold/Rocky \\
\hline
Grass, Bush & High Resistance/Vegetation \\
\hline
\end{tabular}
\caption{Semantic categories and their corresponding terrain type $\tau$.}
\label{tab:terrain_types}
\vspace{-15pt}
\end{table}


\section{Our Approach}

In this section, we discuss how we construct our adaptive multimodal cost map using semantic segmentation and proprioception. 
We highlight the symbols and notation of our approach in table \ref{tab:symbols}.





\subsection{General Knowledge Map} \label{sec:our approach-prior}
Given an RGB image $I_{RGB}^t$, we process it through a semantic segmentation model (\ref{sec:segmentation model-background}) to output a segmented image $I_S^t$. Each pixel in $I_S^t$ categorizes the terrain $\tau$ into the four terrain classes defined in Section \ref{sec:setup-conventions}. Next, we create a \firstmap \, that represents the typical traversability of the terrains ahead of the robot as a costmap. It is constructed by assigning a traversability cost to each segmented terrain class $\tau$ based on the smallest-area ellipse \cite{elnoor2023pronav} corresponding to it. This ellipse represents the distribution of points obtained when using the most stable gait for $\tau$ (see Fig. \ref{fig:ellipses}).

To create \firstmap, we first discretize $I^t_S$ into grids of $n \times n$ pixels as shown in Fig. \ref{fig:cost_maps}.c to obtain $I^t_{SD}$. Each grid in the \firstmap \, is assigned a cost value based on the terrain classes that the pixels in the corresponding grid in $I^t_{SD}$ belong to. Consequently, (\(\mathcal{C}_{\firstletter}^t\)) is constructed as,

\begin{equation}
\mathcal{C}_{\firstletter}^t(i,j) = \frac{\gamma}{n^2} \sum_{(a,b) \in I^t_{SD}(i,j)} \underset{\forall g \in \mathcal{G}}{\mathrm{min}} \big( \mathbf{Ar}(\mathcal{E}^t(\tau_{a,b}, g) \big),
\end{equation}

where, \(C_{\firstletter}^t(i,j)\) denotes the traversability cost for the $(i, j)^{th}$ grid in the \firstmap \, at time t. $\gamma$ is a scaling factor. \(\mathcal{G}\) represents the set of all gaits, and \(\mathbf{Ar}(\mathcal{E}^t(\tau_{a,b}, g))\) refers to the area of the ellipse associated with using gait $g$ in terrain type \(\tau_{a,b}\) present at pixel \((a,b)\).

Additionally, by detecting the predominant terrain $\tau_{pred}$ in a $1 \times 1$ square meter area in front of the robot in $I^t_{SD}$, we can choose the most stable gait to traverse it. This is done based on the gait $g$ that leads to the smallest area for the ellipse \(\mathcal{E}^t(\tau_{pred}, g)\). That is,

\begin{equation}
g^* = \underset{g \in \mathcal{G}}{\mathrm{argmin}}\, \mathbf{Ar} (\mathcal{E}^t(\tau_{pred}, g)).
\end{equation}


\begin{figure}[t]
    \centering
    \includegraphics[width=0.9\columnwidth]{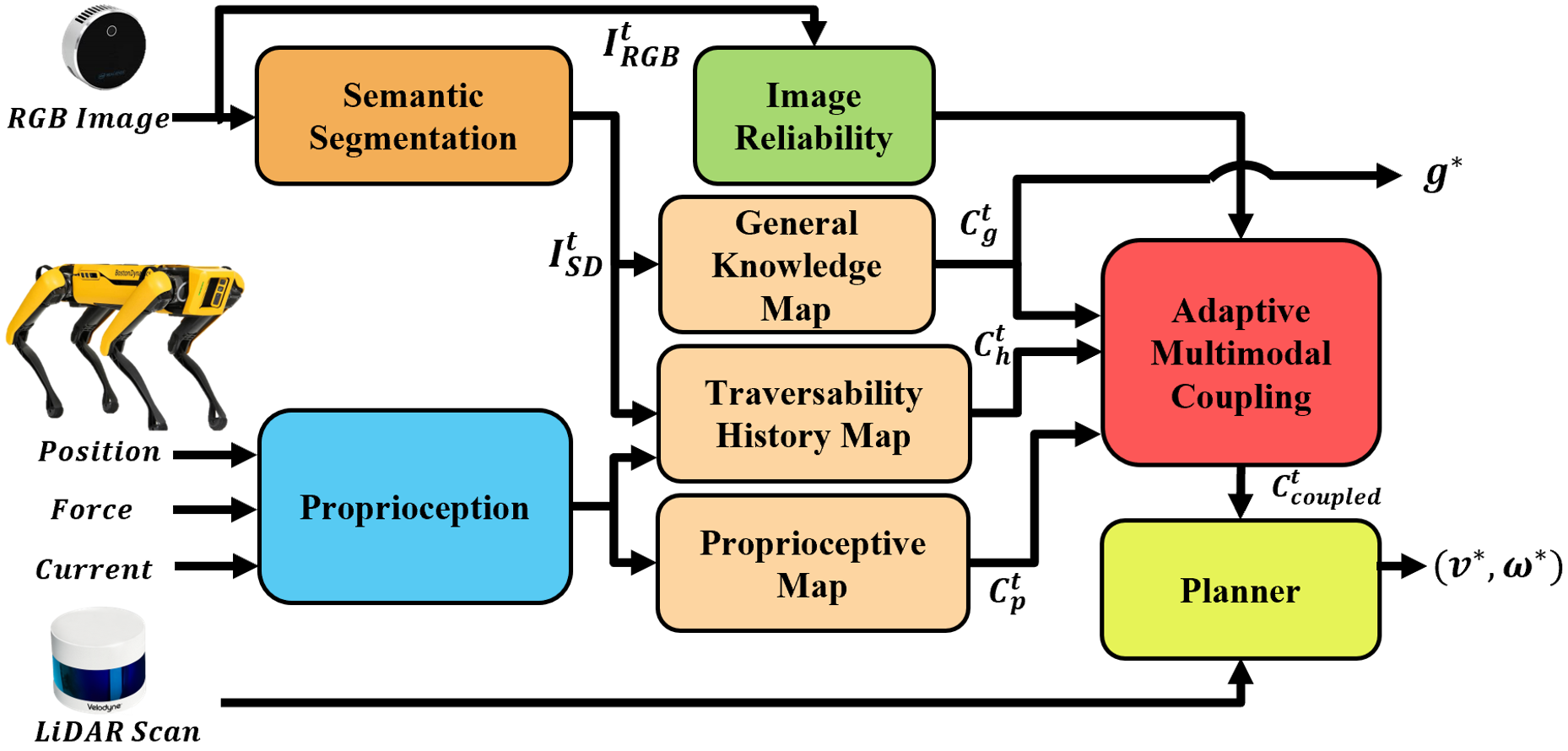}
    \caption{\small{The overall system architecture AMCO that couples vision and proprioception perception modules. We use the robot's joint position, forces, and battery current to compute our  traversability measure. We also use a semantic segmentation module to classify the terrains. After that, three traversability cost maps are created: \firstmap\,  (vision-based), \secondmap\, (vision and proprioception), and \thirdmap\,. All of them are coupled based on the reliability of the input RGB images to create a coupled cost map for the local planner.
   }}
    \label{fig:arch}
    \vspace{-10pt}
\end{figure}


\subsection{Recent Traversability History Map}\label{sec:our approach-history}
We define traversability history of a terrain type $\tau$ as a list of its traversability ($|\mathbf{p}^t_{\tau}|_2$) accumulated over a duration \(T_{\tau}\) by the robot. The intuition behind \(T_{\tau}\) is that the longer the robot has traversed a terrain $\tau$, the more reliable its traversability cost estimate becomes. This history is then coupled with the \firstmap \,  \(C_{\firstletter}^t\). This is because, intuitively, if soil is particularly muddy and unstable on a given day—indicating low stability—it is justified to reflect this observation in the traversability cost for all the soil regions ahead of the robot, even if typically soil is perceived to be more stable by the costs in $\mathcal{C}^t_{\firstletter}$. Building on this intuition, we formulate the recent traversability history as a costmap \(\mathcal{C}_{\secondletter}^t\) for a terrain $\tau$ as, 





\begin{equation}
\mathcal{C}_{\secondletter}^{t}(i,j) =
    \begin{cases}
    \!\begin{aligned}
       & 
        \alpha \cdot \sum_{l=0}^{T_{\tau_{a, b}}} ( \left( 
        \left\lvert\mathbf{p}^{l}_{\tau}\right\rvert_2 -\mathcal{C}_{\firstletter}^{l}(i,j)  \right)), \\
       & \qquad \quad \text{if } (a, b) \in I^t_{SD}(i,j) \,\, \text{and} \,\, \tau_{a, b} \in \Gamma,
    \end{aligned} \\         
    0, \qquad \qquad \text{otherwise}.
  \end{cases}
  \label{eqn:recent-history-map}
\end{equation}

Here, $\Gamma$ is the set of terrains that the robot has traversed in the past, and $T_{\tau_{a, b}}$ is the time the robot spent traversing terrain $\tau_{a, b}$. \(\mathcal{C}_{\secondletter}^{t}(i,j)\) represents the adjusted traversability cost at grid \((i,j)\) and time \(t\), and similarly for \(\mathcal{C}_{\firstletter}^{t}(i,j)\). 
\(\alpha\) is an adjustable factor. Equation \ref{eqn:recent-history-map} ensures that a traversability history cost is applied only to the regions with previously traversed terrains in front of the robot.



\subsection{Reliability Estimation of RGB images} \label{sec:our approach-reliablility}
\begin{figure}[h]
    \centering
    \includegraphics[width=0.8\linewidth]{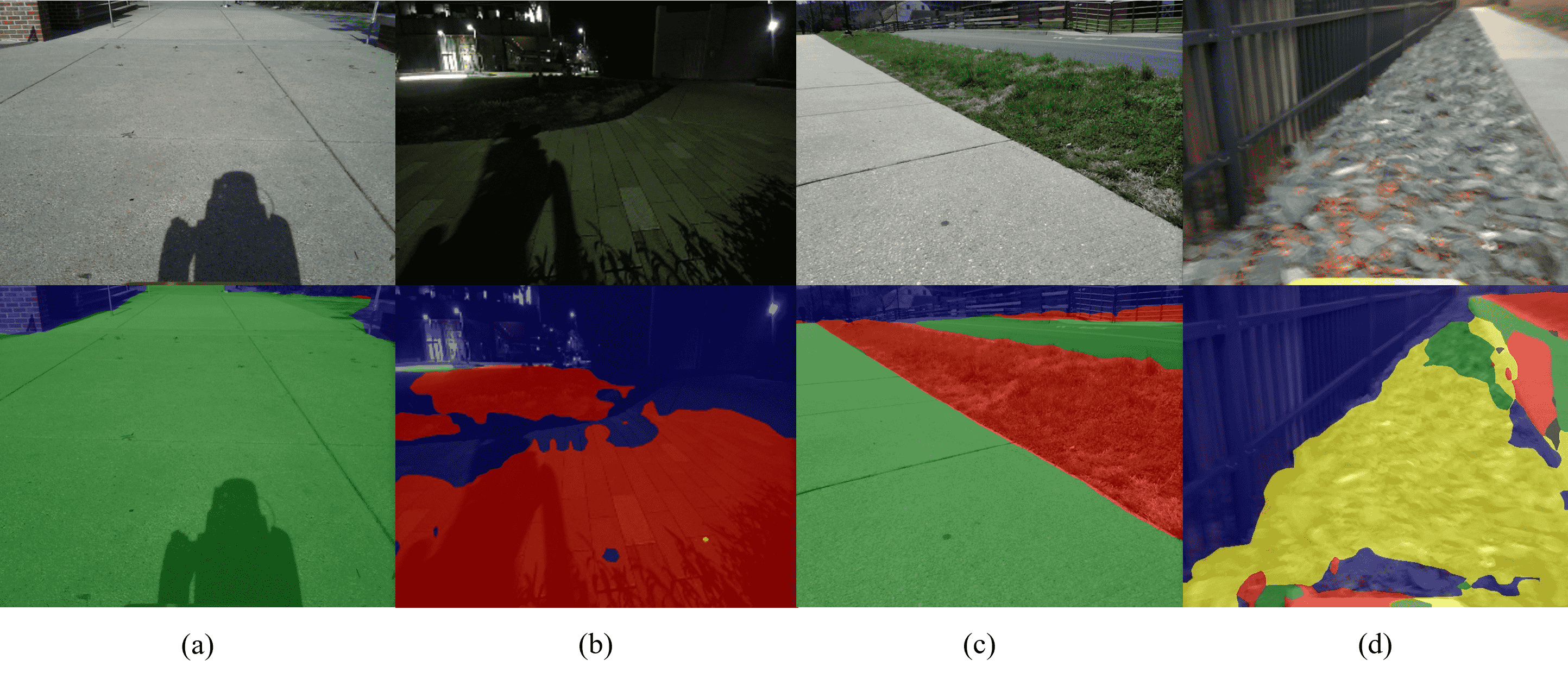}
    \caption{\small{Reliability Estimation of RGB images: we estimate the reliability of RGB images (top row) by measuring two factors, the image's brightness, and blurriness as both affect semantic segmentation (bottom row). (a) and (b) show two brightness levels for an RGB image. (c) shows a sharp image (with almost no blur). (d) shows how rocky terrains could lead to blurriness in the image.}}
    \label{fig:enter-label}
    \vspace{-5pt}
\end{figure}

Assessing the reliability of RGB images is essential for vision-based sensing.
The image's brightness and its blurriness can affect the semantic segmentation (see figure \ref{fig:enter-label}) and consequently the reliability of our \firstmap \,  (\ref{sec:our approach-prior}) and \secondmap\, (\ref{sec:our approach-history}) . Here, we discuss how we evaluate those two visual phenomena:

\subsubsection{Brightness Percentage}
We use two methods to calculate the brightness of an image. The first computes the mean brightness ($r_mean$) of an image as the average of the RGB values for each pixel, normalized to a percentage:\(r_{mean} = \frac{1}{N} \sum_{i=1}^{N} \left( \frac{R_i + G_i + B_i}{3} \right)\),
Where N denotes the number of pixels in the image. \( R_i \), \( G_i \), and \( B_i \) represent the red, green, and blue color values of the \( i^{th} \) pixel, respectively. The arithmetic mean is widely used to estimate the brightness, however, it can be misleading in outdoor environments as it treats all the RGB pixels uniformly. To improve the brightness estimation, we additionally use \cite{bezryadin2007brightness} to compute the luminance (luma) of an image as: \(r_{luma} = \frac{1}{N} \sum_{i=1}^{N} \left
(\rho_1 \cdot R_i + \rho_2 \cdot G_i + \rho_3 \cdot B_i) \right)\).
Similar to $r_{mean}$ but with scalar weights: \(\rho_1 \), \(\rho_2\), \(\rho_3\).

\subsubsection{Blurriness}

An image that is perceived as sharp (not blurry) is typically associated with high fidelity and clear detail representation, which affects semantic segmentation (see figure \ref{fig:enter-label}). To quantify the sharpness/blurriness of an image, we use two methods. The first method employs the Haar wavelet transform \cite{tong2004blur} to derive an energy map $E_{map}$ which can be used to estimate a blurriness score ($r_{hwt}$)of an image. 
Deep-learning methods were used previously 
to detect motion blur \cite{purohit2023learning, nayak2023reinforcement}. However, the datasets for motion blur in outdoor settings are limited. Therefore, we create a dataset with images from unstructured outdoor environments with different degrees of motion blur corruption, and the $\%$ of corruption as their labels. Then, we train a ResNet50 \cite{he2016deep} model on our dataset. 
We represent the output of the model as, $O = \Phi(I^t_{RGB})$. 
We use both methods together to improve blurriness detection.\\
Overall, we formulate the reliability score $\xi$  as follows:
\begin{equation}
\xi = \lambda_{1} \cdot r_{mean} + \lambda_{2} \cdot r_{luma} + \lambda_{3} \cdot r_{hwt} + \lambda_{4} \cdot O
\end{equation}

Here, \(\lambda_{1}\), \(\lambda_{2}\), \(\lambda_{3}\), and \(\lambda_{4}\) are coefficients that balance the contributions of brightness and blur to the overall reliability score. 

\begin{figure}[t]
    \centering
    \includegraphics[width=0.8\linewidth]{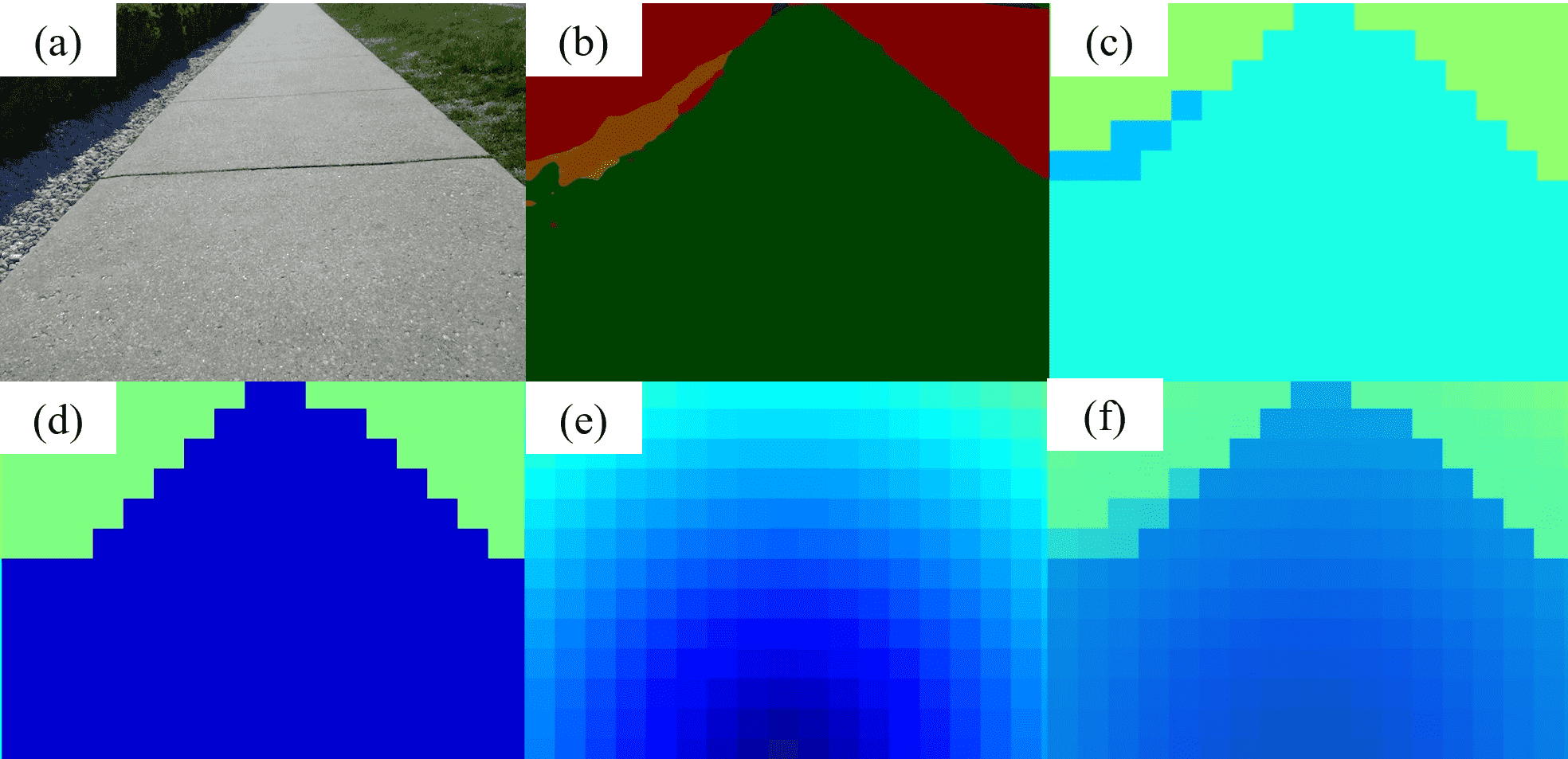}
    \caption{\small{Traversability cost maps of our method. (a) The input RGB image, (b) The segmented image with the color labels as (stable: green, vegetation: red, granular: yellow and poor foothold: orange), (c) The \firstmap\, (vision-based), (d) The \secondmap, (e) The \thirdmap\,. (f) The \finalmap\,. Dark blue reflects low traversability cost in (c-d) maps and conversely red indicates high traversability cost. As shown in the figure, concrete shows low cost in the \firstmap\, and particularly evident in the proprioception-based cost map as the robots captured the terrain stability using proprioception.}}
    \label{fig:cost_maps}
    \vspace{-12pt}
\end{figure}

\subsection{Current Proprioception Map}\label{sec:our approach-proprioception}
Apart from the \firstmap\, and \secondmap\,, we also use the robot's current proprioceptive sensing $|\mathbf{p}^t_{\tau}|_2$ to estimate terrain traversability due to its reliability in indicating how the terrain \textit{feels} regardless of the vision-based sensing. We construct a current \thirdmap\, \(\mathcal{C}_{\thirdletter}\) by extrapolating $|\mathbf{p}^t_{\tau}|_2$ based on the distance \(d_{i,j}\) from the robot's location without using the terrain segmentation (unlike $\mathcal{C}^t_{\firstletter}$ and $\mathcal{C}^t_{\secondletter}$) as, 
\begin{equation}
\mathcal{C}_{\thirdletter}^t(i,j) = \left( \mathcal{U} - \delta \cdot |\mathbf{p}^{t}_{\tau}|_2\right) \cdot (1 - d_{i,j}).
\end{equation}

Here, \(\mathcal{U}\) is a positive value corresponding to the cost of traversing a moderately traversable terrain. We assume a moderately traversable terrain as we do not use the terrain information ahead from the terrain segmentation. \(\mu\) is a scale factor for the traversability's impact on cost. \(d_{i,j}\) is the normalized distance to the \((i,j)^{th}\) grid from the bottom center grid in front of the robot (see Fig. \ref{fig:cost_maps}.e), with values ranging from 0 (directly in front) to 1 (furthest from the robot).

\subsection{Adaptive Multimodal Coupling} \label{sec:adaptive-coupling}
The Adaptive Multimodal Coupling mechanism combines the \firstmap\, $\mathcal{C}_{\firstletter}^t$, \secondmap\, $\mathcal{C}_{\secondletter}^t$, and the \thirdmap\, $\mathcal{C}_{\thirdletter}^t$ into a coupled traversability cost map $\mathcal{C}_{\finalletter}^t$. This integration is achieved through the reliability metric $\xi \in [0-1]$ (Section \ref{sec:our approach-reliablility}), which assesses the quality of RGB images to adaptively adjust the contributions of the vision-based components. The construction of $\mathcal{C}_{\finalletter}^t$ is formulated as,
\begin{equation}
    \mathcal{C}_{\finalletter}^t(i,j) = \xi \cdot (\mathcal{C}_{\firstletter}^t(i,j) + \mathcal{C}_{\secondletter}^t(i,j)) + \mathcal{C}_{\thirdletter}^t(i,j).
\end{equation}

Here, each $(i,j)^{th}$ grid in the input maps results in a corresponding $(i,j)^{th}$ grid in $C_{\finalletter}^t$. The intuition behind this formulation lies in the different reliability of the modalities. The components reliant on vision ($C_{\firstletter}^t$ and $C_{\secondletter}^t$) are susceptible to variations in environmental conditions, such as motion blur, which can diminish their reliability. Conversely, the proprioception-based component $C_{\thirdletter}^t$ provides an accurate measure of the terrain's physical properties, unaffected by visual distortions. Thus, $\xi$  serves as an adaptive scaling factor, reducing the influence of vision-based assessments when their reliability is compromised, and ensuring that \finalmap \,reflects the most accurate traversability information possible.

\begin{figure*}
    \centering
    \includegraphics[width=1.95\columnwidth,height=3.8cm]{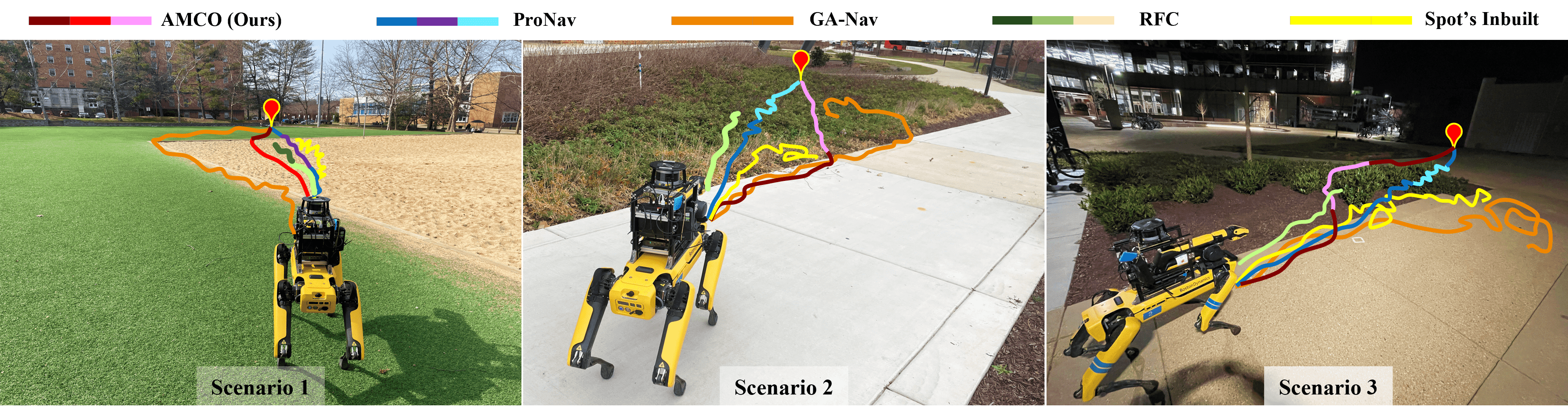}
    \caption{\small{Navigation trajectories of AMCO (Ours) and the other methods in three challenging outdoor scenarios. AMCO, ProNav, and RFC can switch between three gaits (trot, crawl, and amble) during navigation as presented in three different colors.  AMCO, ProNav, and RFC can switch between three gaits (trot, crawl, and amble) during navigation as presented in three different colors. The trajectory colors are represented as follow: AMCO (trot: brown, crawl:red, amble: pink), ProNav \cite{elnoor2023pronav} (trot: blue, crawl:purple, amble: light blue), GaNav \cite{guan2022ga} (orange), Spot's Inbuilt (yellow), and RFC \cite{kertesz2016rigidity} (trot: green, crawl:light green, amble: light beige). Vision-only methods such as GA-Nav fail under low light conditions while proprioception-only methods such as ProNav lead to robot failures in extremely unstable terrains since proprioception cannot estimate trversability beforehand. In contrast, AMCO minimizes the robot's instability during navigation by adaptively coupling both vision and proprioception. }}
    \label{fig:scn-trajs}
    \vspace{-15pt}
\end{figure*}

\subsection{Navigation using the Coupled Traversability Cost Map }


We incorporate a modified version of the Model Predictive Controller (MPC) in \cite{dwa} to generate actions (i.e., ($v,\omega$) linear and angular velocity pairs) for the robot to prefer relatively stable terrains while reaching user-defined local goals. 
MPC in \cite{dwa} calculates a restricted velocity search space $\mathcal{V}_r$ by choosing only the obstacle-free and dynamically feasible actions from an action space $\mathcal{V}_s = \{ (v,\omega) | v \in [0,v_{max}], \omega \in [-\omega_{max},\omega_{max}]\}$, where $v_{max}$ and $\omega_{max}$ are the robot's linear and angular velocity limits.

Then, we define an objective function $J(v,\omega)$ as follows to calculate the optimal action from the restricted velocity space $\mathcal{V}_r$. 

\begin{equation}
\label{eq:dwa_obj_func}
    J(v,\omega) = \beta_1 . goal(v,\omega) + \beta_2 . obs(v,\omega) + \beta_3 . sur(v,\omega) ,
\end{equation}

where $goal(v,\omega)$ and $obs(v,\omega)$ are cost terms that indicate the distance to the goal and proximity to an obstacle for a given action $(v,\omega) \in \mathcal{V}_r$. 
$\beta_i \, \forall i=1,2,3$ are tunable weights. We introduce a traversability cost term $sur(v,\omega)$ to leverage the terrain perception information encoded in the coupled traversability cost map $\mathcal{C}_{\finalletter}$. To calculate the $ sur(v,\omega)$ for a given action $(v,\omega)$, we first extrapolate the action for a $\Delta T$ time and obtain a trajectory $traj^R(v,\omega)$ w.r.t. the robot. Then, $traj^R(v,\omega)$ is projected to the cost map using perspective projection \cite{weerakoon2022graspe}. Let's denote the projected trajectory on the cost map as $traj^C(v,\omega) \in \mathcal{C}_{\finalletter}$. Then, the terrain surface traversability cost $sur(v,\omega)$ is calculated as,

\begin{equation}
    sur(v, \omega) = \frac{1}{|traj^C(v,\omega)|}\sum_{(i,j)\in traj^C(v,\omega)} \mathcal{C}_{\finalletter}^t(i,j),
\end{equation}

where $\mathcal{C}_{\finalletter}(i,j)$ is the traversability cost of the $(i,j)^{th}$ location on the $\mathcal{C}_{\finalletter}^t$ at time t, and $|traj^C(v,\omega)|$ is the cardinality of the $traj^C(v,\omega)$. Finally, the optimal navigation action $(v^*,\omega^*)$ is obtained as,

\begin{equation}
    (v^*,\omega^*) = \underset{(v,\omega) \in \mathcal{V}_r}{\operatorname{argmin}} J(v,\omega).
\end{equation}

After that, we send $(v^*,\omega^*)$ along with $g^*$ (which is computed from equation 2) to the robot's controller.

\section{Results and Analysis}

\subsection{Implmentation \& Robot Setup}
We implement AMCO on Boston Dynamics Spot, a quadruped robot with 12
degrees of freedom (DOF) which provides joint feedback (position, velocities, and forces) from the robot legs and the current consumed by the battery. Additionally, the robot is equipped with a VLP16 Velodyne lidar and a Zed 2i camera. AMCO runs on an Intel NUC edge computer equipped with an Intel i7 CPU, and an Nvidia RTX2060 GPU. 
We use the following other parameters for our implementation: $n = 40, \gamma = 1,  \alpha = 4.5 \rho_1 = 0.299, \rho_2 = 0.587, \rho_3 = 0.114, \delta = \, 31.875, \mathcal{U}\, = 127,  \lambda_1 =0.0008, \lambda_2 = 0.001, \lambda_3 = \, 0.0025, \lambda_4 = \, 0.003,
  v_{max} = \, 0.6 \, m/s, \omega_{max} = \, 0.2 \, rad/s, \beta_1 =\,2.4 , \beta_2 \, = \,3.2 , \beta_3 \,= 0.05$.

\subsection{Comparison Methods}

\begin{itemize}
    \item \textbf{Spot's in-built planner}:  A planner that uses RGB-D cameras to detect and avoid obstacles. 


    \item \textbf{GA-Nav} \cite{guan2022ga}: A vision-based method that uses semantic segmentation of RGB images to compute traversability costs for different terrains. It is combined with a local planner \cite{weerakoon2022terp} to compute trajectories on low-cost terrains and avoid obstacles. 
    
    \item \textbf{ProNav} \cite{elnoor2023pronav}: A proprioceptive-based method that uses joint feedback and battery's current to estimate traversability and changes to an appropriate gait while computing goal-conditioned velocities using \cite{sathyamoorthy2023vern}.

    \item \textbf{Random Forest Classifier (RFC)} \cite{kertesz2016rigidity}: It uses proprioceptive signals to classify the terrain type. We used the dataset in \cite{elnoor2023pronav} to train the classifier to identify between four different terrain types (granular, poor foothold, high resistance and stable). RFC is combined with \cite{sathyamoorthy2023vern} to generate goal-directed velocities $(v^*, \omega^*)$.

\end{itemize}

\subsection{Evaluation Metrics}\label{sec:metrics}
\begin{itemize}
    \item \textbf{Success Rate}: The ratio of successful navigation trials where the robot was able to reach its goal without freezing or colliding with obstacles.

    \item \textbf{Normalized Trajectory Length:} The ratio between the robot’s trajectory length and the straight-line distance to the goal averaged across both successful and unsuccessful trials.

    \item \textbf{IMU Energy Density}: The mean of the aggregated squared acceleration values measured by the IMU sensors across the x, y, and z axes, calculated over the successful trials \cite{try2023vibration}: \( E_i = \sum_{n=1}^{N} a_i^2 \), and \( E_{\text{Total}} = E_{ax} + E_{ay} + E_{az} \), where \( a_i \) represents one of the three acceleration signals (x, y, and z axes), and \( N \) is the IMU readings along the trajectory.
    
   \item \textbf{Vibration Cost}: 
    The cumulative sum of deviations, measured from a predefined stable reference terrain, in the actuator positions at the robot's hip joints along its trajectory \cite{elnoor2023pronav}:
    {\small
    \[
    \delta_{j}(t) = 
    \begin{cases}
        \left| p^{j}(t) - \min(p^{j}_{\text{ref}}) \right| & \text{if } p^{j}(t) < \min(p^{j}_{\text{ref}}), \\
        \left| p^{j}(t) - \max(p^{j}_{\text{ref}}) \right| & \text{if } p^{j}(t) > \max(p^{j}_{\text{ref}}). \\
    \end{cases}
    \]
    }
    Where, \( p^{j}(t) \) is the position of joint \( j \) at time \( t \), and \( p^{j}_{\text{ref}} \) represents reference positions from a stable terrain (concrete). The total Vibration Cost at time \( t \) is then computed as \( \text{Vibration Cost}(t) = \sum_{j} \delta_{j}(t) \).
    

\end{itemize}

\subsection{Test Scenarios}
\begin{itemize}
    \item \textbf{Scenario 1:} contains granular terrain (sand) and grass. 
    \item \textbf{Scenario 2:} contains dense vegetation.
    \item \textbf{Scenario 3:} contains bushes, mulch, and concrete. 
    \item \textbf{Scenario 4:} contains mud, grass and concrete.
\end{itemize}

\begin{table}[t]

\centering
\scriptsize
\begin{tabularx}{\columnwidth}{|c|c|X|X|X|X|}
\hline
\textbf{Metrics} & \textbf{Method} & \textbf{Scen. 1} & \textbf{Scen. 2} & \textbf{Scen. 3} & 
\textbf{Scen. 4}\\
\hline
\multirow{6}{*}{\shortstack{Success\\Rate (\%) $\uparrow$}} 
& In-built system   & 10 &  10& - & -  \\
& GA-Nav\cite{guan2022ga}    & 80 & 30& 20 &60\\
& RFC\cite{kertesz2016rigidity}    & 40 & 50& 60 & 30\\
& ProNav\cite{elnoor2023pronav}   & 60  & 60& 70 & 40\\
& AMCO w/o reliability &70 &  60 & 50 &80 \\
& AMCO w/o history  & 70 &   70& 80 &60\\
& AMCO (ours)   & \textbf{100} &   \textbf{90}& \textbf{100} & \textbf{90}\\
\hline
\multirow{6}{*}{\shortstack{Normalized \\Trajectory\\ Length}} 
& In-built system  & 1.215 &   1.420 & - &- \\
& GA-Nav\cite{guan2022ga}   & 1.324 & 1.394& 2.105 &1.447  \\
& RFC\cite{kertesz2016rigidity}   & 1.089& 0.916& 1.129& 0.838 \\
& ProNav\cite{elnoor2023pronav}     &1.063  & 0.948& 1.036&0.861 \\
& AMCO w/o reliability  &1.192 & 1.131 &1.235 &1.217\\
& AMCO w/o history & 1.230  & 1.178& 1.189& 1.253\\
& AMCO (ours)    & 1.171 & 1.064& 1.093&1.184  \\
\hline
\multirow{6}{*}{\shortstack{IMU\\Energy Density $\downarrow$}} 
& In-built system  & 37506 & 51764& -&  - \\
& GA-Nav\cite{guan2022ga}     & 18392 & 29354& 21349 & 26096\\
& RFC\cite{kertesz2016rigidity}   & 23412& 30233 & 26043 & 37412 \\
& ProNav\cite{elnoor2023pronav}  & 21342 &   26490 & 24392 & 35291\\
& AMCO w/o reliability   & 14323& 25374 & 25034 &24523 \\
& AMCO w/o history   & 16271& 24361& 22375 &25828 \\
& AMCO (ours)     & \textbf{13928} & \textbf{19832}& \textbf{20126} &\textbf{23120} \\
\hline
\multirow{6}{*}{\shortstack{Vibration\\Cost  $\downarrow$}} 
& In-built system  & 29.3  & 48.7 & - & - \\
& GA-Nav\cite{guan2022ga}    & 14.3  & 28.5& 23.9 & 19.4\\
& RFC\cite{kertesz2016rigidity}  & 15.9  & 29.2& 37.4 &36.7 \\
& ProNav\cite{elnoor2023pronav}   & 16.4  & 31.3 & 34.9 &  34.1 \\
& AMCO w/o reliability & 12.3 & 24.1 & 26.6& 19.1\\
& AMCO w/o history & 11.5 & 22.8& 21.4& 23.4 \\
& AMCO (ours) & \textbf{9.3}  &  \textbf{20.3} & \textbf{19.3} &\textbf{17.3} \\
\hline

\end{tabularx}
\caption{\small{The table shows the quantitative results of AMCO and the other methods over 10 trials. We use four navigation metrics to evaluate our method's performance. All metrics are averaged over both the successful and unsuccessful trails (reaching the goal).}}
\label{tab:tab2}
\vspace{-15pt}
\end{table}

\subsection{Analysis and Discussion}
 We evaluated our method's navigation performance qualitatively in Figs. \ref{fig:cover} and \ref{fig:scn-trajs},  and quantitatively in Table \ref{tab:tab2}.  
 Across the four scenarios, our method AMCO consistently outperforms the other methods in terms of successful goal-reaching by effectively selecting appropriate gaits and velocities to reach the designated goal. Further, AMCO results in lower vibration experienced by the robot in all scenarios compared to the other methods.
 
\no \textbf{In Scenario 1}, GA-Nav generates significantly longer trajectories preferring grass over the granular sand terrain, and demonstrates oscillatory behavior near the boundary between the two terrains.
ProNav and RFC use relatively straight trajectories and switch to crawl gait after encountering heavy instability in the granular sand region. However, in certain instances, their inconsistent gait selection leads to sinkage in the sand. In contrast, AMCO switches to the crawl gait in advance using the general knowledge map to minimize the potential instabilities in the forthcoming terrain (i.e., sand). 

\no \textbf{Scenario 2} depicts a navigation task on concrete followed by a densely vegetated region. ProNav leads to leg entanglement in vegetation due to its delayed gait switching. GANav's perception of vegetation is affected by image motion blur which leads to inconsistent segmentation. However, AMCO generates trajectories to maximize the stay in the concrete region while switching to the amble gait before entering the vegetated region to reduce the leg entanglement risks.

\no \textbf{In Scenario 3}, our method AMCO relies heavily on \thirdmap \,for navigation due to low and varying lighting conditions which results in less reliable RGB images. Both Spot's in-built system and GANav struggles to reach the goal due to erroneous perception from the camera images under challenging lighting. The complexity of the muddy terrain in \textbf{Scenario 4} poses significant challenges to all the methods. Particularly,  RFC and ProNav methods experience considerable leg sinkage due to delayed gait switching. Although GA-Nav attempts to take longer trajectories through grass and concrete, the method struggles in muddy regions due to its fixed gait. In contrast, AMCO initially adjusts its gait using the general knowledge map and later changes the trajectory to avoid the heavily deformable muddy regions to reach the goal.


\no \textbf{Benefits of Image Reliability:} We observe that our method's navigation performance is significantly affected without the image reliability estimation, especially under low-light conditions (scenario 3) and motion blur (scenarios 2 and 4). This is primarily due to the inconsistent segmentation impacts on the general knowledge map, which eventually results in errors in the coupled traversability cost estimation. However, AMCO's adaptive coupling which accounts for image reliability ensures prioritizing the proprioception under unreliable image inputs.

\no \textbf{Benefits of Traversability History:} We observe a notable decrease in success rate when traversability history is omitted from AMCO's formulation. Particularly in scenarios 1 and 4, where the visually similar-looking granular terrain's deformability varies from one region to another. However, the use of traversability history allows AMCO to avoid severely deformable regions based on recent experiences on the same terrain. Further, the recent history enables AMCO to update the forthcoming terrain conditions better, compared to relying only on fixed costs provided by image segmentation.

\section{Conclusion, Limitations \& Future work}

We present AMCO, a novel approach for quadruped robot navigation through challenging outdoor environments by adaptively combining vision-based and proprioceptive-based perception capabilities. Our method effectively utilizes semantic segmentation from RGB images and proprioceptive inputs, to generate a coupled perception representation. This enables the selection of appropriate gaits and velocities, which significantly improve navigation success rates and stability across various terrains.
Our method has some limitations. For instance, AMCO's reliance on proprioceptive data under dark conditions limits its capabilities. It's a good direction to explore integrating other sensor modalities in evaluating traversability such as (thermal or hyperspectral cameras). Furthermore, incorporating Vision Language Models (VLMs) could offer more insights into complex environments, due to their common reasoning capabilities.


\bibliographystyle{IEEEtran}
\bibliography{references}

\end{document}